\documentclass[letterpaper]{article} 
\usepackage[submission]{aaai2027}  
\usepackage[hyphens]{url}  
\usepackage{graphicx} 
\usepackage{natbib}  
\usepackage{caption} 
\usepackage{algorithm}
\usepackage{algorithmic}
\usepackage{amsfonts} 
\usepackage{amsmath}
\usepackage{multirow} 
\usepackage{pifont}
\newcommand{\cmark}{\ding{51}}
\newcommand{\xmark}{\ding{55}}

\usepackage{newfloat}
\usepackage{listings}
\DeclareCaptionStyle{ruled}{labelfont=normalfont,labelsep=colon,strut=off} 
\floatstyle{ruled}
\newfloat{listing}{tb}{lst}{}
\floatname{listing}{Listing}

\usepackage{booktabs}

\title{Semantic Evidence Regulation via Relational Bias for Zero-Shot Object Navigation}
\author{
    Weitao An\textsuperscript{\rm 1},
    Chenghao Xu\textsuperscript{\rm 2},
    Xu Yang\textsuperscript{\rm 1},
    Cheng Deng\textsuperscript{\rm 2}
}
\affiliations{
    \textsuperscript{\rm 1}School of Electronic Engineering, Xidian University, Xi'an 710071, China\\
    \textsuperscript{\rm 2}School of Information Science and Engineering, Hohai University, Nanjing 210098, China\\

}

\begin{document}

\maketitle

\begin{abstract}

Object navigation requires an embodied agent to locate a target object in an unknown environment through visual observations. Existing zero-shot methods typically leverage open-vocabulary perception and semantic priors to identify promising search regions. However, these methods often assume that semantic cues provide reliable guidance, while lacking mechanisms to assess and adjust their influence during exploration. Consequently, misleading semantic evidence can persist throughout navigation, biasing frontier selection and causing agents to repeatedly explore unreliable regions.
To address this problem, we propose \textbf{SER-Nav}, a training-free framework that achieves dynamic \textbf{S}emantic \textbf{E}vidence \textbf{R}egulation via relational bias. SER-Nav introduces dual relational biases: activation reinforces target-related and contextual evidence to guide exploration, while inhibition attenuates misleading evidence caused by perceptual confusion and failed verification. These relational biases dynamically update the navigation search space, and a reliability-aware commitment gate prevents premature pursuit of insufficiently verified targets. By adaptively adjusting the influence of semantic cues, SER-Nav preserves useful guidance while mitigating persistent semantic misguidance under noisy open-vocabulary perception. 
Experiments on standard ObjectNav benchmarks demonstrate that SER-Nav consistently improves success rate and path efficiency over representative zero-shot methods. Extensive ablation studies further validate the effectiveness of each component. Real-world robot experiments further validate its robustness and practical applicability.

\end{abstract}


\section{Introduction}

\begin{figure}[t]
    \centering
    \includegraphics[width=\linewidth]{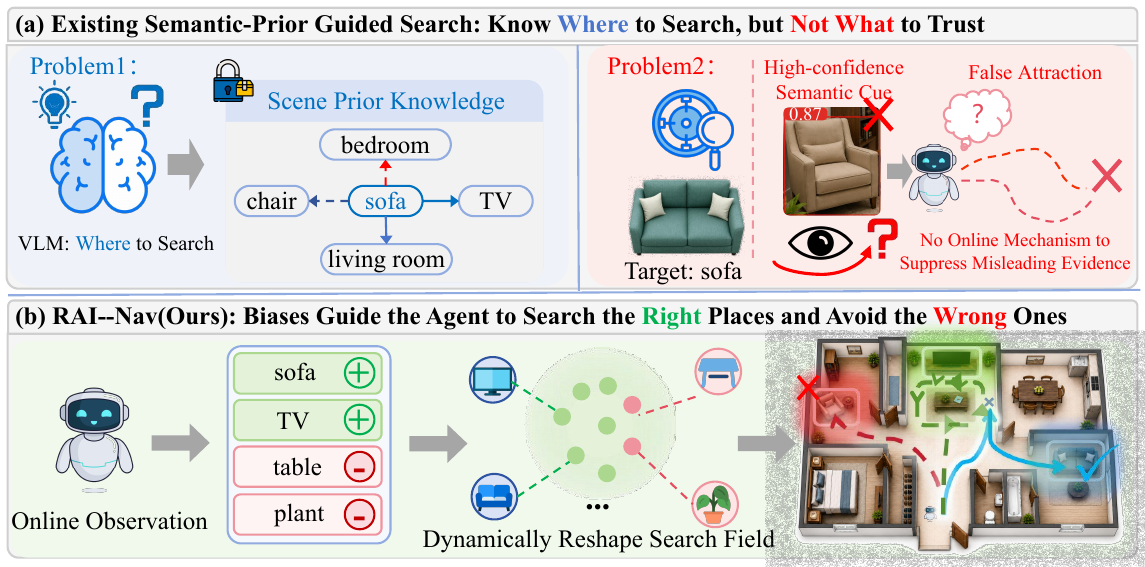}
    \caption{
    \textbf{Motivation of SER-Nav.} (a) Existing semantic-prior guided search determines \emph{where to search} based on static priors, but can be misled by unreliable semantic cues and distractors.
    (b) SER-Nav models dual relational biases, with positive activation and negative inhibition, to reshape the search field and guide the agent toward reliable regions.
    }
    \label{fig:motivation}
\end{figure}

Embodied object navigation requires an agent to explore an unknown
environment and reach a target object category~\cite{batra2020objectnav}.
Beyond spatial mapping and path planning, the agent must interpret target
semantics, scene context, and uncertain perceptual
information~\cite{savva2019habitat,chang2017matterport3d,ramakrishnan2021hm3d}.
In indoor environments, targets are often occluded, small, distant, or
visually similar to distractors, making open-world object search
particularly challenging.

Existing approaches broadly follow three strategies. Learning-based
policies directly map observations and goals to actions, but often require
large-scale training and remain sensitive to environment-specific
distributions~\cite{gupta2017cognitive,wijmans2020ddppo,ye2021auxiliary}.
Map-driven methods construct occupancy, semantic, object, or scene
representations for frontier selection and target
search~\cite{chaplot2020topological,chaplot2020semexp,georgakis2021learning,procthor2022,l3mvn2023,pixnav2024},
but their decisions are tightly coupled with perception outputs.
Egocentric online methods use first-person observations and
vision-language reasoning for flexible
planning~\cite{yokoyama2024vlfm,instructnav2024,trihelper2024,imaginenav2025,unigoal2025,mfnp2025,ascent2025,cos2026,panonav2026},
yet remain sensitive to local views, prompt design, and unstable model
responses. Despite their differences, these methods commonly rely on
semantic observations, learned priors, or commonsense relations to
determine where to search.

Despite its utility in guiding frontier selection, semantic guidance is not always reliable under noisy open-vocabulary perception. As illustrated
in Fig.~\ref{fig:motivation}, false detections and category confusion can
create misleading search cues, while missed detections may obscure
promising regions. Moreover, static semantic priors and accumulated map
evidence are difficult to revise once they influence frontier selection,
even when contradicted by subsequent observations or embodied
interactions. Consequently, erroneous or outdated cues may persist across
navigation steps and repeatedly direct the agent toward unproductive
regions. Robust ObjectNav should therefore not only infer
\emph{where to search}, but also determine \emph{what not to trust} and
revise semantic guidance when previous hypotheses are contradicted.

In real-world environments, semantic cues are inherently relational rather than independent~\cite{yin2024sgnav}. For example, when searching for a sofa, co-occurring objects such as TVs or coffee tables provide supportive contextual evidence, whereas visually similar objects such as chairs may introduce ambiguous cues. Beyond these semantic relations, failed access attempts provide online feedback that previously promising regions may be unreliable. These relations naturally provide both positive and negative evidence, suggesting that semantic guidance should be dynamically strengthened or weakened according to online observations rather than treated as fixed positive evidence.

Motivated by such insight, we propose \textbf{SER-Nav}, a training-free
navigation framework that dynamically regulates semantic evidence 
through dual relational biases of activation and inhibition. SER-Nav first aggregates multi-view
observations into object-level relational evidence, reducing the influence
of individual noisy detections. It then propagates activation from target
and contextual evidence to promote promising frontiers, while propagating
inhibition from perceptually confusing objects and failed access events to
suppress misleading or falsified regions. These signals are integrated in
a dynamic Relational Activation--Inhibition Exploration Graph that updates
frontier values as new evidence arrives, while a reliability-aware
commitment gate prevents premature pursuit of insufficiently verified
targets. By adjusting the influence of semantic guidance through online
observation and embodied feedback, SER-Nav mitigates persistent semantic
misguidance and enables more reliable and efficient exploration without
online LLM reasoning.

In summary, our contributions are threefold:
\begin{itemize}
    \item We introduce a relational-bias perspective, reformulating embodied navigation as a search-space modulation problem jointly governed by activation and inhibition biases.

    \item We propose SER-Nav, a training-free relational activation–inhibition exploration framework that dynamically propagates supportive and contradictory evidence to promote reliable frontiers, suppress misleading regions, and regulate target commitment. 

    \item Extensive experiments on standard ObjectNav benchmarks and a
    real-world robot demonstrate that SER-Nav improves success rate and
    path efficiency while remaining robust under noisy
    open-vocabulary perception.
\end{itemize}

\section{Related Work}

\begin{figure*}[t]
    \centering
    \includegraphics[width=0.98\textwidth]{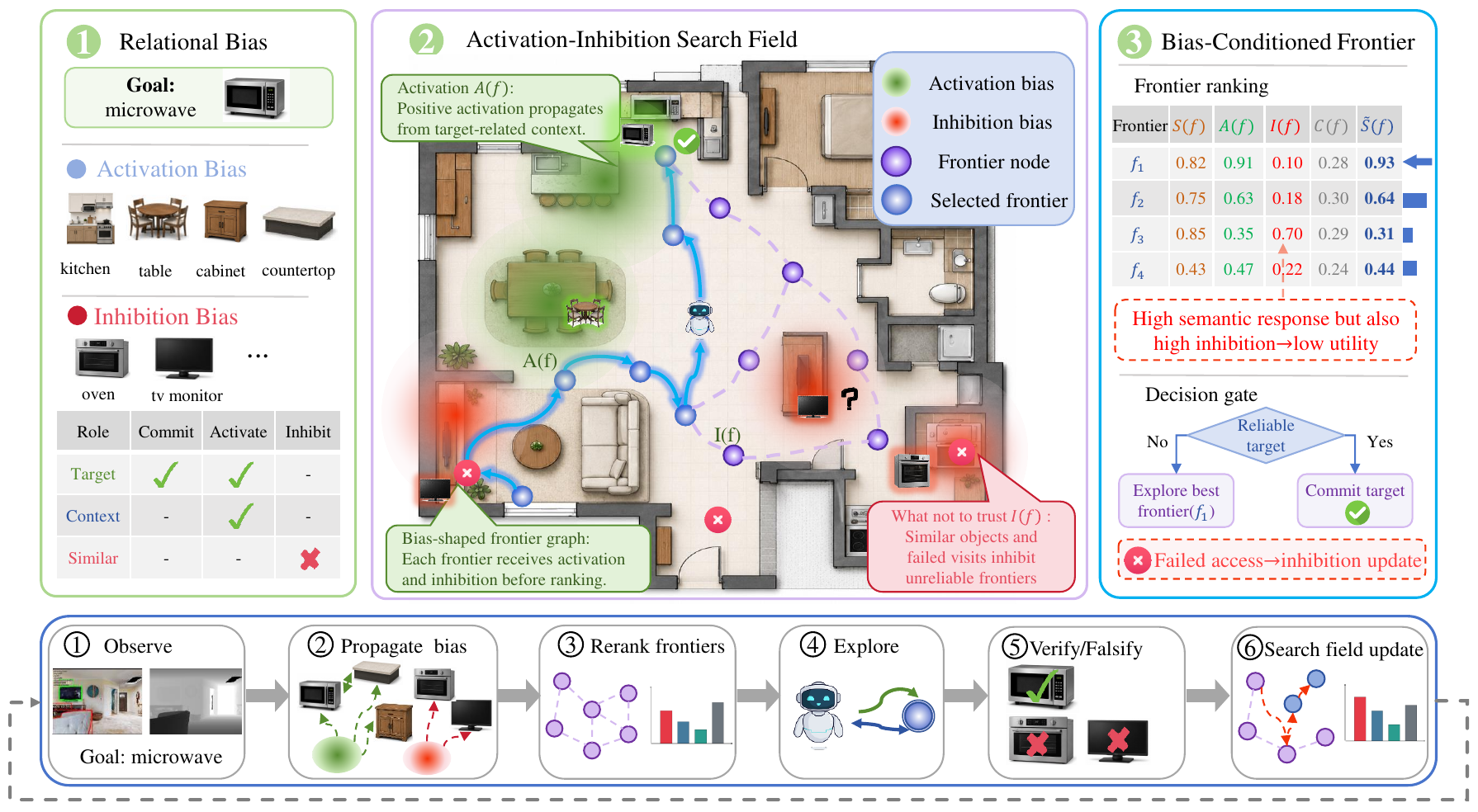}
    \caption{
    \textbf{Overview of SER-Nav.} Given a target category, SER-Nav first constructs target-centric category-role vectors that assign semantic cues to commitment, activation, and inhibition roles. Online object observations are converted into object-level relational evidence through multi-view accumulation and multi-label object-role competition. Positive affinity cues generate activation over promising frontiers, while similar distractors and failed access events generate inhibition over unreliable regions. The agent finally selects frontiers through bias-conditioned score updating and commits to the target only when object-level evidence is sufficiently reliable.
    }
    \label{fig:method_overview}
\end{figure*}

\subsection{Object Navigation with Semantic Mapping}
Object navigation has been extensively studied in Habitat, Matterport3D, Gibson, AI2-THOR, RoboTHOR, and HM3D~\cite{savva2019habitat,chang2017matterport3d,xia2018gibson,kolve2017ai2thor,deitke2020robothor,ramakrishnan2021hm3d}, and the ObjectNav benchmark standardizes evaluation of agents that locate target objects in unseen environments~\cite{batra2020objectnav}. Methods range from learned visual policies to modular mapping-and-planning~\cite{gupta2017cognitive,wijmans2020ddppo,ye2021auxiliary}, often relying on explicit spatial or semantic memories including frontier exploration, Neural SLAM, topological and semantic maps~\cite{yamauchi1997frontier,chaplot2020active,chaplot2020topological,georgakis2021learning}. Representative methods such as Goal-Oriented Semantic Exploration and PONI~\cite{chaplot2020semexp,ramakrishnan2022poni} use semantic maps to select long-term goals, but typically treat semantic cues only as positive evidence. SER-Nav instead modulates frontier selection when cues can be both informative and misleading.

\subsection{Structured Scene Priors for Navigation}
Structured scene priors infer likely target locations beyond geometry. Scene priors, occupancy anticipation, 3D scene graphs, open-vocabulary 3D representations, and object-centric maps have been used for navigation~\cite{yang2019scenepriors,ramakrishnan2020occupancy,armeni2019scenegraph,peng2023openscene,gu2024conceptgraphs,werby2024hovsg}. Recent methods transfer structured scene representations to zero-shot navigation via soft constraints, scene-graph prompting, Voronoi-based topology, and dynamic object-relation graphs~\cite{zhou2023esc,yin2024sgnav,wu2024voronav,tang2024openobjectnav}, but mostly use relations as positive priors. SER-Nav factorizes target-centric relations into dual roles: affinity activates promising frontiers, while inhibition suppresses unreliable regions caused by visual confusion or failed verification.

\subsection{Open-Vocabulary Object Navigation}
Open-vocabulary ObjectNav leverages vision-language models (CLIP, BLIP-2~\cite{radford2021clip,li2023blip2}) and open-vocabulary detectors (GroundingDINO, Segment Anything, MobileSAM, YOLO-World, OWL-ViT~\cite{liu2023groundingdino,kirillov2023sam,zhang2023mobilesam,cheng2024yoloworld,minderer2022owlvit}) to extend navigation beyond fixed categories\cite{li2018self,deng2018triplet,gong2019differentiable}. Methods such as ZSON, CoW, VLFM, OpenFMNav, OVExp, and ApexNav~\cite{majumdar2022zson,gadre2023cow,yokoyama2024vlfm,kuang2024openfmnav,wei2024ovexp,zhang2025apexnav} enable zero-shot search or adaptive exploration, but remain vulnerable to false positives and unverified contextual cues. SER-Nav complements these by keeping the perception stack while adding a lightweight relational decision layer that converts semantic observations into activation and inhibition evidence.

\section{Method}

\begin{figure}[t]
    \centering
    \includegraphics[width=\columnwidth]{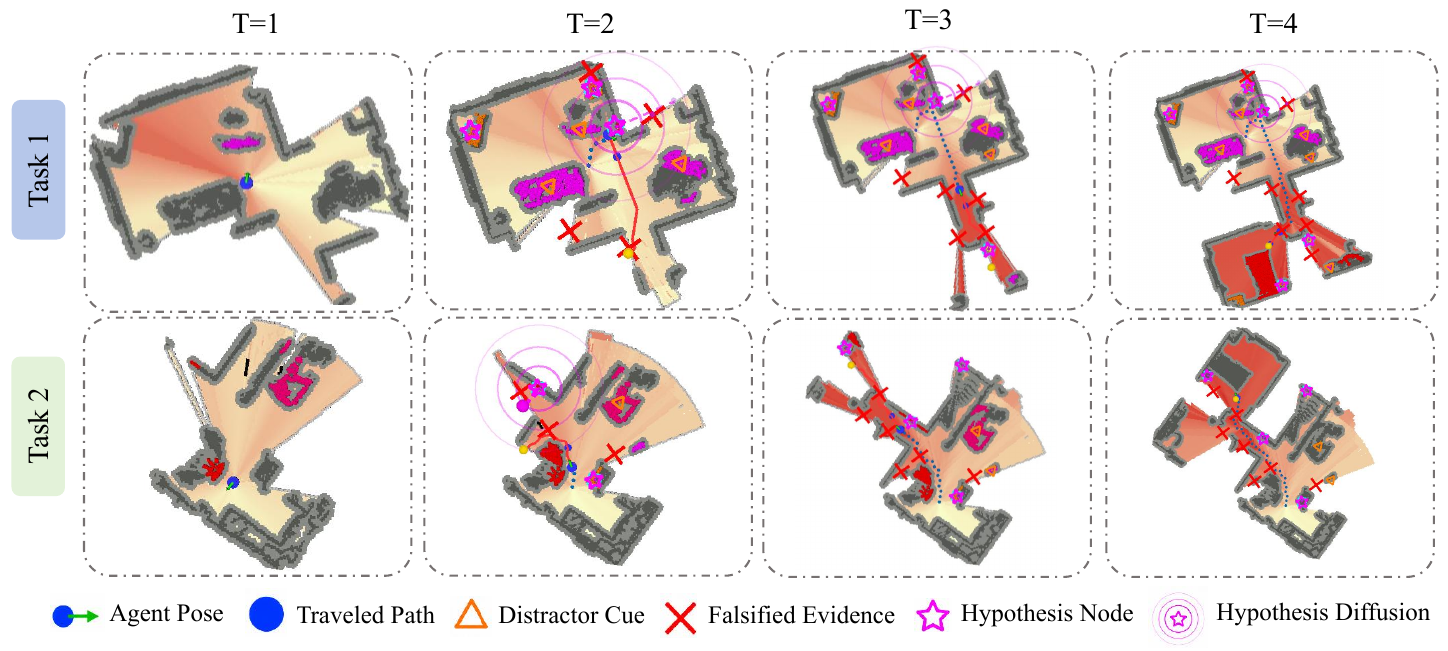}
    \caption{
    \textbf{Temporal evolution of relational activation--inhibition propagation.} Contextual observations generate hypothesis nodes and diffusion fields, while distractor cues and falsified evidence suppress unreliable regions. As exploration proceeds, SER-Nav dynamically updates the search field for subsequent frontier selection.
    }
    \label{fig:propagation_detail}
\end{figure}

\begin{table*}[t]
\centering
\caption{\textbf{Comparison with representative ObjectNav methods.} TF denotes training-free navigation. ZS denotes zero-shot setting. LLM denotes online LLM reasoning.}
\label{tab:main_results}
\begin{tabular}{l c c c cc cc cc}
\toprule
\multirow{2}{*}{Method} 
& \multirow{2}{*}{TF} 
& \multirow{2}{*}{ZS} 
& \multirow{2}{*}{LLM}
& \multicolumn{2}{c}{HM3Dv2} 
& \multicolumn{2}{c}{HM3Dv1} 
& \multicolumn{2}{c}{MP3D} \\
\cmidrule(lr){5-6} \cmidrule(lr){7-8} \cmidrule(lr){9-10}
& & & & SR$\uparrow$ & SPL$\uparrow$ 
& SR$\uparrow$ & SPL$\uparrow$ 
& SR$\uparrow$ & SPL$\uparrow$ \\
\midrule
PONI~\cite{ramakrishnan2022poni}        & \xmark & \xmark & \xmark & --   & --   & --   & --   & 31.8 & 12.1 \\
ProcTHOR~\cite{procthor2022}            & \xmark & \xmark & \xmark & --   & --   & 54.4 & 31.8 & --   & --   \\
ZSON~\cite{majumdar2022zson}            & \xmark & \xmark & \xmark & --   & --   & 25.5 & 12.6 & 15.3 & 4.8  \\
CoW~\cite{gadre2023cow}                 & \cmark & \cmark & \xmark & --   & --   & 32.0 & 18.1 & 9.2  & 4.9  \\
ESC~\cite{zhou2023esc}                  & \cmark & \cmark & \cmark & --   & --   & 39.2 & 22.3 & 28.7 & 14.2 \\
L3MVN~\cite{l3mvn2023}                  & \cmark & \cmark & \cmark & 36.3 & 15.7 & 50.4 & 23.1 & 34.9 & 14.5 \\
PixNav~\cite{pixnav2024}                & \cmark & \cmark & \cmark & --   & --   & 37.9 & 20.5 & --   & --   \\
VoroNav~\cite{wu2024voronav}            & \cmark & \cmark & \cmark & --   & --   & 42.0 & 26.0 & --   & --   \\
InstructNav~\cite{instructnav2024}      & \cmark & \cmark & \cmark & 58.0 & 20.9 & 58.0 & 20.9 & --   & --   \\
VLFM~\cite{yokoyama2024vlfm}            & \cmark & \cmark & \xmark & 63.6 & 32.5 & 52.5 & 30.4 & 36.4 & 17.5 \\
TriHelper~\cite{trihelper2024}          & \cmark & \cmark & \cmark & --   & --   & 56.5 & 25.3 & --   & --   \\
SG-Nav~\cite{yin2024sgnav}              & \cmark & \cmark & \cmark & 49.6 & 25.5 & 54.0 & 24.9 & 40.2 & 16.0 \\
ImagineNav~\cite{imaginenav2025}        & \cmark & \cmark & \cmark & --   & --   & 53.0 & 23.8 & --   & --   \\
UniGoal~\cite{unigoal2025}              & \cmark & \cmark & \cmark & --   & --   & 54.0 & 24.9 & 41.0 & 16.4 \\
MFNP~\cite{mfnp2025}                    & \cmark & \cmark & \cmark & --   & --   & 58.3 & 26.7 & 41.1 & 15.4 \\
CoS~\cite{cos2026}                      & \cmark & \cmark & \xmark & --   & --   & 55.9 & 29.1 & 37.6 & 17.6 \\
PanoNav~\cite{panonav2026}              & \cmark & \cmark & \cmark & --   & --   & 43.5 & 23.7 & --   & --   \\
\midrule
\textbf{SER-Nav} & \cmark & \cmark & \xmark
& \textbf{74.5} & \textbf{36.8} 
& \textbf{60.5} & \textbf{32.1} 
& \textbf{42.1}& \textbf{19.2} \\
\bottomrule
\end{tabular}
\end{table*}

\paragraph{Task Definition.}
We consider \textit{zero-shot 3D indoor object navigation}, where an agent receives first-person observations $o_t$ and a target category $g$ at each discrete time step $t$, and succeeds only if it executes \textsc{Stop} within a predefined success radius of the target object before reaching the maximum step budget.

\paragraph{Method Overview.}
As shown in Fig.~\ref{fig:method_overview}, SER-Nav reshapes frontier exploration through target-centric dual relational biases, aiming to dynamically regulate the influence of reliable and unreliable semantic evidence. Given a target category, we first assign open-vocabulary semantic cues to three action roles: target commitment, positive activation, and negative inhibition. Online observations are then accumulated into object-level relational evidence through multi-view label competition. Finally, SER-Nav propagates activation from target evidence as well as contextual evidence, and propagates inhibition from visually similar distractors or failed access events. It then updates frontier scores with these dual biases, and switches to target commitment only when the object-level evidence is sufficiently reliable.

\subsection{Target-centric Relational Bias Construction}

Given a target category $g$, SER-Nav first constructs a target-centric relational bias that converts open-vocabulary perception outputs from category labels into action-aware relational evidence. Specifically, each target category is associated with a target-specific relation set consisting of contextual and visually similar categories. These relations are encoded as a three-dimensional category-role vector for each category $\ell$:
\begin{equation}
\mathbf{b}_g(\ell) =
\left[
c_g(\ell), a_g(\ell), i_g(\ell)
\right]^\top
\in \{0,1\}^3,
\label{eq:category_role_vector}
\end{equation}
where $c_g(\ell)$ indicates whether category $\ell$ provides target-commitment evidence, $a_g(\ell)$ indicates whether it provides positive search activation, and $i_g(\ell)$ indicates whether it provides negative inhibition. Target categories and their synonyms are encoded as $[1,1,0]^\top$, contextual co-occurrence categories as $[0,1,0]^\top$, and visually similar distractors as $[0,0,1]^\top$. 

The online perception module produces object observations:
\begin{equation}
d_i = (\mathbf{p}_i, s_i, \ell_i),
\end{equation}
where $\mathbf{p}_i \in \mathbb{R}^2$ is the projected map position, $s_i \in [0,1]$ is the detection confidence, and $\ell_i$ is the detected category. These observations provide the spatial, confidence, and semantic information required to build object-level relational evidence~\cite{yang2021r3det}.

Since open-vocabulary detections are noisy, the same physical object may be assigned different labels across views~\cite{yang2019deep}. SER-Nav therefore maintains each object cluster as a multi-label entity instead of fixing its semantic role from a single observation. For each object cluster $o$ and candidate label $\ell \in \mathcal{L}(o)$, we estimate a label reliability score:
\begin{equation}
\rho_o(\ell)
=
m_o(\ell)\, r_o(\ell),
\label{eq:label_reliability}
\end{equation}
where $m_o(\ell)$ denotes the accumulated multi-view support for semantic hypothesis $\ell$ of object $o$, and $r_o(\ell)$ denotes the aggregated confidence reliability of this hypothesis. The dominant semantic hypothesis is then selected through multi-label competition:
\begin{equation}
\ell_o^{\star}
=
\arg\max_{\ell \in \mathcal{L}(o)}
\rho_o(\ell).
\label{eq:dominant_label}
\end{equation}

The reliability of the selected label is further converted into an object-level evidence strength:
\begin{equation}
r_o
=
\sigma
\left(
\rho_o(\ell_o^{\star})
\right),
\label{eq:object_strength}
\end{equation}
where $\sigma(\cdot)$ is a bounded normalization function. Finally, the object-level relational evidence is obtained by assigning the dominant label to its target-centric action role:
\begin{equation}
\mathbf{h}_o
=
r_o \mathbf{b}_g(\ell_o^{\star})
=
\left[
h^c_o,
h^a_o,
h^i_o
\right]^\top.
\label{eq:object_evidence}
\end{equation}
Here, $h^c_o$, $h^a_o$, and $h^i_o$ denote target-commitment evidence, positive activation evidence, and negative inhibition evidence, respectively. This multi-label object-role competition makes the object role depend on accumulated multi-view evidence rather than a single noisy detection, while exposing an action-aware role profile for downstream activation--inhibition propagation.

\subsection{Relational Activation--Inhibition Propagation}

Based on the object-level relational evidence, SER-Nav propagates local object observations into the frontier decision space. At decision step $t$, we construct a local relational graph:
\begin{equation}
\mathcal{G}_t = (\mathcal{V}_t, \mathcal{E}_t),
\end{equation}
where $\mathcal{V}_t$ contains object evidence nodes $\mathcal{O}_t$, contextual hypothesis nodes $\mathcal{H}_t$, negative evidence nodes $\mathcal{Y}_t$, and candidate frontiers $\mathcal{F}_t$. The edge set $\mathcal{E}_t$ encodes local spatial adjacency among nodes, such that relational evidence is propagated only within a local neighborhood. Each node $v \in \mathcal{V}_t$ has a 2D map position $\mathbf{p}_v$.

We define the spatial propagation kernel as:
\begin{equation}
\kappa(\mathbf{p}_v, \mathbf{p}_u)
=
\exp\left(
-\frac{\|\mathbf{p}_v-\mathbf{p}_u\|_2}{R}
\right)
\mathbb{I}
\left[
\|\mathbf{p}_v-\mathbf{p}_u\|_2 \leq R
\right],
\label{eq:spatial_kernel}
\end{equation}
where $R$ is the propagation radius and $\mathbb{I}[\cdot]$ is the indicator function. This kernel implements distance-decayed propagation over local graph neighborhoods.

For a candidate frontier $f \in \mathcal{F}_t$, its relational activation is defined as:
\begin{equation}
A_t(f)
=
\sum_{o \in \mathcal{O}_t}
h^c_{o,t}
\kappa(\mathbf{p}_f,\mathbf{p}_{o})
+
\sum_{q \in \mathcal{H}_t}
a_{q,t}
\kappa(\mathbf{p}_f,\mathbf{p}_{q}),
\label{eq:activation}
\end{equation}
where $h^c_{o,t}$ denotes the target-commitment evidence of object $o$, and $\mathcal{H}_t=\Psi_h(\mathcal{O}_t^{\mathrm{ctx}})$ denotes the set of contextual hypothesis nodes generated from contextual objects $\mathcal{O}_t^{\mathrm{ctx}}=\{o\in\mathcal{O}_t\mid c_g(\ell_o^\star)=0,\ a_g(\ell_o^\star)=1\}$. For each hypothesis node $q\in\mathcal{H}_t$, $\mathbf{p}_q$ and $a_{q,t}$ denote its position and intrinsic activation strength, respectively (see Supplementary Section A.4). The first term propagates direct target or synonym evidence from observed objects, while the second term expands context-induced hypotheses to nearby candidate frontiers. 

Similarly, the relational inhibition of frontier $f$ is defined as:
\begin{equation}
I_t(f)
=
\sum_{o \in \mathcal{O}_t}
h^i_{o,t}
\kappa(\mathbf{p}_f,\mathbf{p}_{o})
+
\sum_{n \in \mathcal{Y}_t}
a_{n,t}
\kappa(\mathbf{p}_f,\mathbf{p}_{n}),
\label{eq:inhibition}
\end{equation}
where $h^i_{o,t}$ is the inhibition evidence mainly induced by visually similar distractors, and $a_{n,t}$ denotes the intrinsic inhibition strength of negative-evidence node $n$ (see Supplementary Section A.4). Failed access events are incorporated into $\mathcal{Y}_t$ when a frontier is visited without obtaining a reliable target candidate. In this case, SER-Nav generates or strengthens a negative-evidence node at that position, which temporarily reduces the priority of nearby frontiers. The inhibition strength decays over time and can be overridden by new consistent evidence, allowing recovery from transient perception failures and preventing excessive suppression of valuable exploration regions. Thus, contextual cues are treated as verifiable hypotheses rather than static positive priors.

In summary, $A_t(f)$ encodes \emph{where to search}, while $I_t(f)$ encodes \emph{where not to trust}. Through graph-based activation--inhibition propagation, SER-Nav dynamically reshapes the frontier search space with both positive and negative relational cues. Figure~\ref{fig:propagation_detail} visualizes this temporal update process.

\subsection{Bias-conditioned Exploration and Target Commitment}

Given the candidate frontier set $\mathcal{F}_t$, SER-Nav updates the original semantic response of each frontier through dual-bias modulation. Let $S^0_t(f)$ denote the base semantic response of frontier $f$, which is computed using BLIP-2~\cite{li2023blip2} to measure the semantic alignment between the target category and detected object regions, and let $\widetilde{S}_{t}(f)$ denote its bias-conditioned updated score. We compute:
\begin{equation}
\widetilde{S}_{t}(f)
=
\frac{
S^0_t(f) + A_t(f)
}{
1 +  I_{t}(f) + \bar{C}_t(f)
},
\label{eq:score_update}
\end{equation}
where $A_t(f)$ and $I_{t}(f)$ are the activation and inhibition terms, and $\bar{C}_t(f)$ is the normalized path cost from the current agent position to frontier $f$. This update increases the priority of frontiers supported by target-related semantic and contextual evidence, while suppressing frontiers associated with distractors, falsified regions, repeated visits, or high navigation cost.

The next exploration frontier is selected by:
\begin{equation}
f_t^{\star}
=
\arg\max_{f \in \mathcal{F}_t}
\widetilde{S}_t(f).
\label{eq:frontier_selection}
\end{equation}

In addition to frontier exploration, SER-Nav uses the commitment channel to decide when to switch from exploration to target commitment. Specifically, we define:
\begin{equation}
o_t^{\star}
=
\arg\max_{o \in \mathcal{O}_t}
h^c_{o,t},
\qquad
\Gamma_t
=
\mathbb{I}
\left[
h^c_{o_t^{\star},t}
\geq
\theta
\right],
\label{eq:commit_gate}
\end{equation}
where $h^c_{o,t}$ is the target-commitment evidence of object $o$, $\theta$ is the commitment threshold, and $\Gamma_t$ is the commitment gate. The final decision at step $t$ is:
\begin{equation}
\pi_t =
\begin{cases}
\pi_{\mathrm{commit}}(o_t^{\star}), & \Gamma_t = 1, \\
\pi_{\mathrm{explore}}(f_t^{\star}), & \Gamma_t = 0.
\end{cases}
\label{eq:policy}
\end{equation}

Thus, SER-Nav performs bias-conditioned exploration before the target is reliably verified, and switches to target commitment only when object-level evidence is sufficiently strong. Once committed, SER-Nav uses an A$^\ast$ local planner to approach the selected target candidate through step-wise discrete actions, avoiding premature stopping and repeated exploration of unreliable regions.

\section{Experiment}

\begin{figure}[t]
    \centering
    \includegraphics[width=\linewidth]{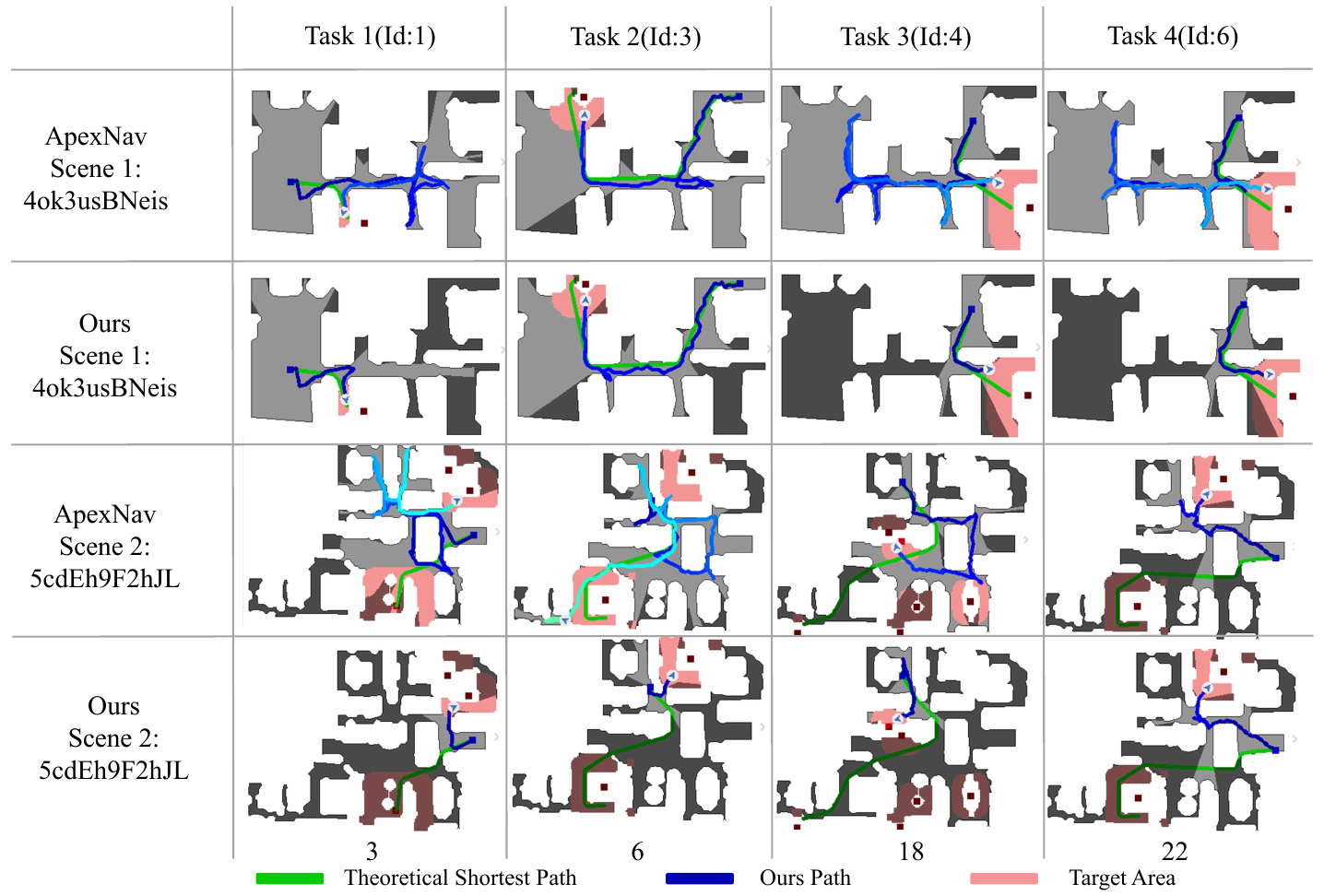}
    \caption{
    \textbf{Trajectory comparison between the baseline method and ours on representative ObjectNav tasks.} Our method reduces redundant exploration and selects more reliable search directions under noisy semantic observations.
    }
    \label{fig:apexnav_comparison}
\end{figure}

\subsection{Experiment Setup}

\paragraph{Datasets.}
We evaluate SER-Nav in the Habitat simulator on three standard ObjectNav benchmarks: HM3Dv1, HM3Dv2, and MP3D~\cite{ramakrishnan2021hm3d,yadav2022hm3dsem,chang2017matterport3d}. HM3Dv1 contains 2,000 episodes across 20 scenes and 6 goal categories. HM3Dv2 contains 1,000 episodes across 36 scenes and 6 goal categories. MP3D contains 2,195 episodes across 11 scenes and 21 goal categories.

\paragraph{Implementation details.}
Following the evaluation settings of ApexNav~\cite{zhang2025apexnav}, the agent acts in a discrete action space, including \textsc{MoveForward} $(0.25\mathrm{m})$, \textsc{TurnLeft} $(30^\circ)$, \textsc{TurnRight} $(30^\circ)$, and \textsc{Stop}, with a maximum budget of 500 steps per episode. The agent operates within a perception range of $[0.0\mathrm{m}, 5.0\mathrm{m}]$, and success is determined by the Habitat evaluator's $0.2\mathrm{m}$ valid-\textsc{Stop} criterion. To isolate the effect of executive control, all variants adopt the same perception--mapping--planning stack, using YOLO, GroundingDINO, MobileSAM, and BLIP-2. All experiments are run on a single NVIDIA RTX A6000 GPU with approximately 8GB of memory.

\paragraph{Evaluation metrics.}
We evaluate navigation performance using Success Rate (SR) and Success weighted by Path Length (SPL). All results are reported in percentages. For ablation analysis, we additionally report SoftSPL to measure partial progress toward the goal when an episode fails. Detailed definitions of all evaluation metrics are provided in the supplementary material.

\subsection{Main Results}

\paragraph{Comparison with representative ObjectNav methods.} Table~\ref{tab:main_results} compares SER-Nav with representative ObjectNav methods across HM3Dv2, HM3Dv1, and MP3D. SER-Nav achieves the better performance on HM3Dv2 and HM3Dv1, obtaining 74.5 SR and 36.8 SPL on HM3Dv2, and 60.5 SR and 32.1 SPL on HM3Dv1. Compared with VLFM, which is also training-free, zero-shot, and does not rely on online LLM reasoning, SER-Nav improves SR by 10.9 points and SPL by 4.3 points on HM3Dv2. On HM3Dv1, SER-Nav also outperforms recent zero-shot and LLM-based methods, showing that relational activation--inhibition improves both search success and path efficiency. 
On MP3D, SER-Nav achieves competitive performance without online LLM reasoning. 
Its lightweight relational decision layer provides a favorable trade-off between accuracy, robustness, and inference cost under noisy open-vocabulary perception.

\begin{figure}[t]
    \centering
    \includegraphics[width=\linewidth]{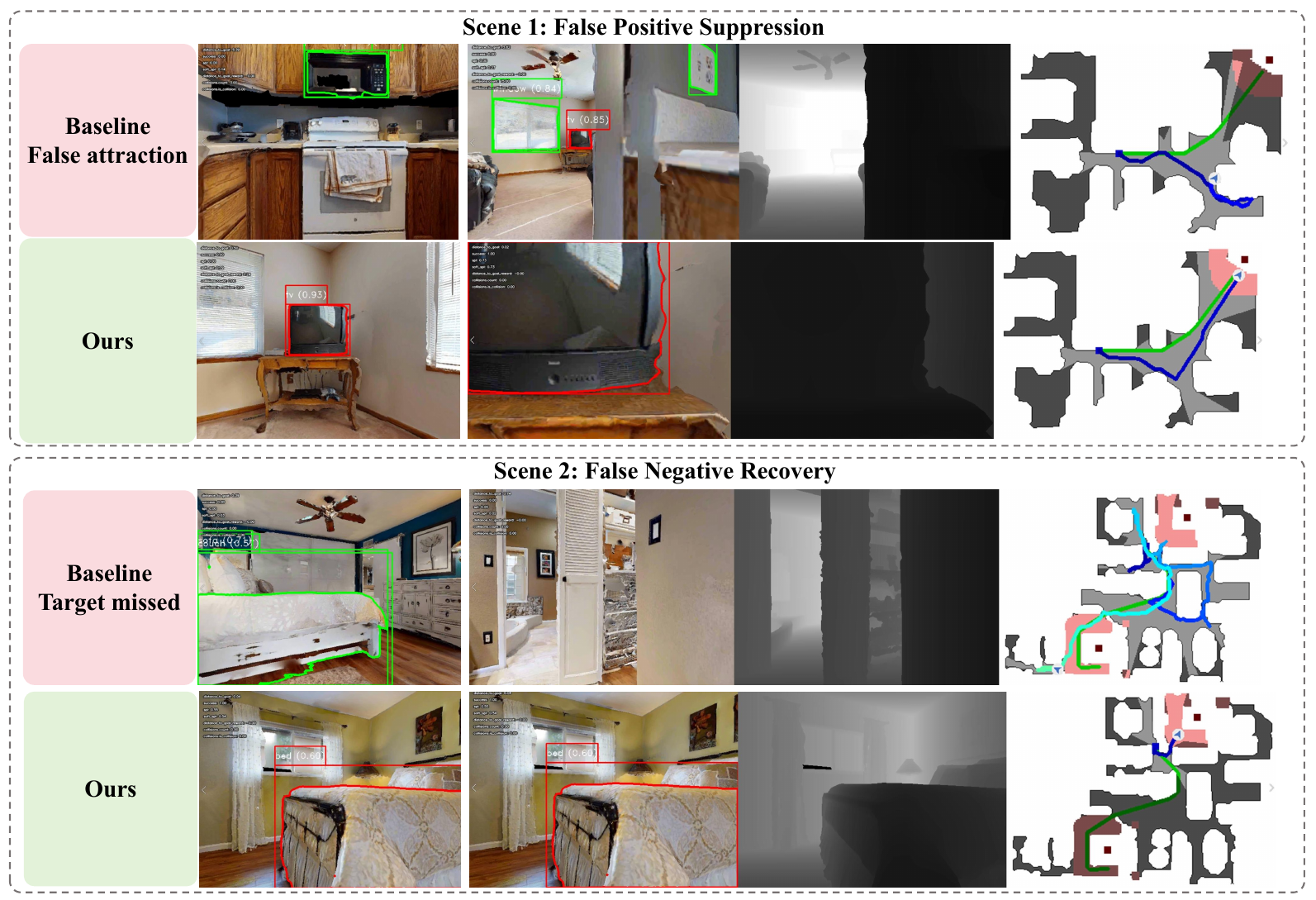}
    \caption{
    \textbf{Qualitative analysis under noisy open-vocabulary perception.}
    In Scene 1, high-confidence false detections attract the baseline toward unreliable regions, while SER-Nav suppresses false-positive attraction. In Scene 2, the baseline misses the target under unreliable observations, whereas SER-Nav recovers through contextual activation and action-level verification.
    }
    \label{fig:false_positive_negative}
\end{figure}

\paragraph{Qualitative analysis.} Figure~\ref{fig:apexnav_comparison} and Figure~\ref{fig:false_positive_negative} provide qualitative evidence for the proposed dual-bias mechanism. In noisy open-vocabulary perception, the baseline can be misled by high-confidence false detections or miss the target under unreliable observations. SER-Nav mitigates these failures by suppressing false-positive regions through inhibition evidence and recovering from weak target evidence through contextual activation and action-level verification. As a result, SER-Nav produces more reliable trajectories, reduces redundant exploration, and selects search directions that better align with the target area.

\subsection{Ablation Analysis}

\begin{table}[t]
\centering
\caption{\textbf{Ablation study of SER-Nav variants on the HM3Dv2 validation set.}}
\begin{tabular}{lccc}
\toprule
Method & SR $\uparrow$ & SPL $\uparrow$ & SoftSPL $\uparrow$ \\
\midrule
w/o Activation Bias & 65.4 & 29.2 & 32.0 \\
w/o Similar Inhibition & 69.8 & 33.0 & 35.7 \\
w/o Failed Access & 71.3 & 32.7 & 35.5 \\
w/o Multi-label Competition & 66.7 & 30.4 & 33.3 \\
\textbf{SER-Nav Full} & \textbf{74.5} & \textbf{36.8} & \textbf{38.9} \\
\bottomrule
\end{tabular}
\label{tab:ablation_hm3dv2}
\end{table}

\paragraph{Ablation study.} Table~\ref{tab:ablation_hm3dv2} evaluates the contribution of each core component on the HM3Dv2 validation set. Removing the activation bias causes a substantial drop from 74.5 to 65.4 SR, demonstrating the importance of positive contextual affinity for identifying promising search regions. Removing similar-object inhibition also degrades performance, reducing SR to 69.8 and SPL to 33.0, confirming that visually similar distractors can mislead open-vocabulary navigation. Without failed-access memory, SR drops to 71.3, indicating that action-level falsification helps avoid repeated exploration of already verified but unproductive regions. Removing multi-label competition reduces SR to 66.7 and SPL to 30.4, showing that multi-view label evidence is important for robust object-role assignment under noisy detections. 

\begin{table}[t]
\centering
\caption{\textbf{Effect of object detectors on navigation performance (HM3Dv1 val).} G-DINO represents GroundingDINO. Each method is shown with two detector variants.}
\setlength{\tabcolsep}{3.2pt} 
\begin{tabular}{l l c c c}
\toprule
Method & Detector & SR $\uparrow$ & SPL $\uparrow$ & SoftSPL $\uparrow$ \\
\midrule
VLFM & G-DINO & 45.6 & 26.1 & - \\
     & YOLO + G-DINO & 52.5 & 30.4 & - \\
\midrule
ApexNav & G-DINO & 43.7 & 23.7 & 26.4 \\
        & YOLO + G-DINO & 52.8 & 27.1 & 30.1 \\
\midrule
\textbf{SER-Nav} & \textbf{G-DINO} & \textbf{57.3} & \textbf{29.4} & \textbf{32.9} \\
       & \textbf{YOLO + G-DINO} & \textbf{60.5} & \textbf{32.1} & \textbf{34.5} \\
\bottomrule
\end{tabular}
\label{tab:detector_effect_grouped}
\end{table}

\paragraph{Effect of object detectors.} Table~\ref{tab:detector_effect_grouped} studies the impact of perception quality on navigation performance. Replacing G-DINO with the stronger YOLO+G-DINO perception stack improves all methods, confirming that better open-vocabulary detection generally benefits ObjectNav. However, SER-Nav consistently outperforms VLFM and ApexNav under both detector settings. With G-DINO alone, SER-Nav achieves 57.3 SR, outperforming VLFM by 11.7 points and ApexNav by 13.6 points. With YOLO+G-DINO, SER-Nav further improves to 60.5 SR and 32.1 SPL. These results suggest that SER-Nav does not merely benefit from stronger detectors, but more effectively converts noisy semantic outputs into reliable activation and inhibition evidence. 



\subsection{Efficiency \& Sensitivity Analysis}

\begin{table}[t]
\centering
\caption{\textbf{Efficiency comparison on HM3D.} LLM Calls denotes the average number of online LLM calls per episode. Runtime is measured in seconds per episode.}
\label{tab:efficiency}
\setlength{\tabcolsep}{3.5pt}
\begin{tabular}{lcccc}
\toprule
Method & SR$\uparrow$ & LLM Calls$\downarrow$ & Steps$\downarrow$ & Runtime (s)$\downarrow$ \\
\midrule
L3MVN        & 51.5 & 35.0  & 194.1 & 1047.7 \\
PixNav      & 48.2 & 58.1  & 278.7 & 276.4  \\
SG-Nav       & 54.2 & 122.3 & 300.0 & 728.5  \\
InstructNav & 54.4 & 113.6 & 320.5 & 458.8  \\
ASCENT       & 57.6 & 2.0   & 171.4 & 190.3  \\
\midrule
\textbf{SER-Nav} & \textbf{60.5} & \textbf{0.0} & \textbf{144.0} & \textbf{101.8} \\
\bottomrule
\end{tabular}
\end{table}

\paragraph{Efficiency comparison.} Table~\ref{tab:efficiency} compares SER-Nav with representative LLM or VLM-based ObjectNav methods in terms of online reasoning cost and runtime. Many recent methods require frequent online LLM calls, leading to high runtime per episode. For example, SG-Nav and InstructNav require more than 100 online LLM calls per episode on average. In contrast, SER-Nav does not rely on online LLM reasoning. It achieves the best SR among the compared methods while using fewer navigation steps and lower runtime. This demonstrates that the proposed relation-based decision layer is both effective and lightweight.

\begin{figure}[t]
    \centering
    \includegraphics[width=\linewidth]{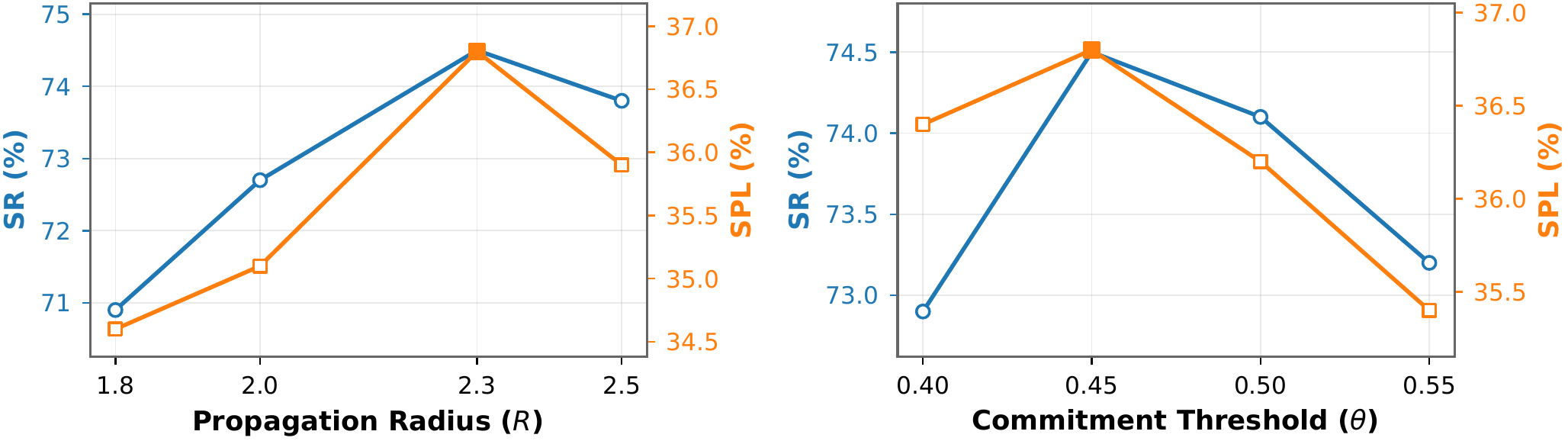}
    \caption{\textbf{Hyperparameter sensitivity analysis on the HM3Dv2 validation set.} SER-Nav achieves the best overall performance at $R=2.3$ and $\theta=0.45$.
    }
    \label{fig:sensitivity_r_theta}
\end{figure}

\paragraph{Hyperparameter sensitivity.} Figure~\ref{fig:sensitivity_r_theta} reports the sensitivity of SER-Nav to the propagation radius $R$ and the commitment threshold $\theta$ on the HM3Dv2 validation set. SER-Nav remains stable within a reasonable range of hyperparameter values, and the best overall performance is achieved at $R=2.3$ and $\theta=0.45$, which are used as the default settings in our experiments. 

\begin{table}[t]
\centering
\caption{\textbf{Relation source analysis on HM3Dv2.} LLM Relation reports the mean and standard deviation over three independently generated relation sets.}
\label{tab:relation_source}
\begin{tabular}{lccc}
\toprule
Relation Source & SR$\uparrow$ & SPL$\uparrow$ & SoftSPL$\uparrow$ \\
\midrule
Random Relation  & 61.2 & 27.1 & 28.3 \\
LLM Relation     & $73.6{\pm}1.4$ & $36.2{\pm}0.9$ & $38.3{\pm}1.1$ \\
\midrule
\textbf{Default Relation} & \textbf{74.5} & \textbf{36.8} & \textbf{38.9} \\
\bottomrule
\end{tabular}
\end{table}

Table~\ref{tab:relation_source} analyzes the impact of relation construction on navigation performance. Random relations lead to degraded performance, achieving 61.2 SR, 27.1 SPL, and 28.3 SoftSPL, which indicates that arbitrary semantic associations may introduce unreliable relational biases for navigation. In contrast, relation sets generated by the LLM achieve consistently better performance, with an average of $73.6{\pm}1.4$ SR, $36.2{\pm}0.9$ SPL, and $38.3{\pm}1.1$ SoftSPL over three independently generated sets. These results demonstrate that meaningful relational biases, including both supportive and inhibitory relations, play an important role in effective semantic evidence regulation. The default relation set was generated once by DeepSeek before evaluation, without validation-based selection or manual modification.

\subsection{Real-world Robot Validation}
\begin{figure}[t]
    \centering
    \includegraphics[width=\columnwidth]{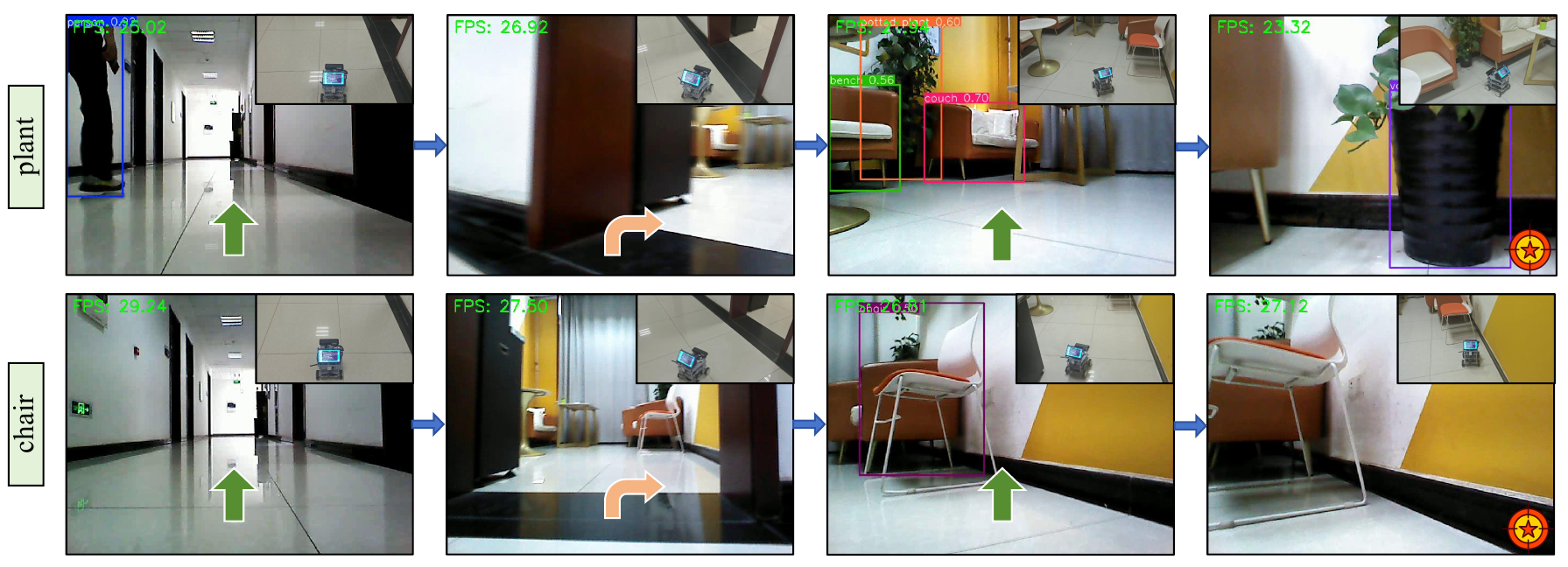}
    \caption{
    \textbf{Real-world validation of SER-Nav.}
    Two representative object navigation tasks are shown in a real indoor environment. Without additional tuning, SER-Nav successfully localizes and reaches the target objects under practical sensing noise, demonstrating effective sim-to-real transfer.
    }
    \label{fig:real_world}
\end{figure}

To validate the practical applicability of SER-Nav, we deploy the proposed framework on a Wheeltec R550 mobile robot equipped with an Astra Pro Plus RGB-D camera and an NVIDIA Jetson Orin NX Super 16GB. Open-vocabulary perception is executed on a remote workstation with an NVIDIA RTX A6000 GPU, while localization, mapping, and motion control run onboard. The robot and server communicate through ROS~2 over a local network.

Figure~\ref{fig:real_world} shows two representative navigation tasks in a real indoor environment. Without any task-specific tuning, SER-Nav successfully locates and reaches the target objects by leveraging contextual activation while suppressing misleading semantic observations. These results demonstrate that the proposed dual-bias exploration strategy transfers effectively from simulation to real-world deployment.

\section{Conclusion}

We presented \textbf{SER-Nav}, a zero-shot embodied navigation framework that uses dual relational biases to dynamically guide exploration in unknown environments. By factorizing target-centric object relations into positive activation and negative inhibition,  it modulates frontier selection and commits to targets only when object-level evidence is reliable. Extensive experiments on standard ObjectNav benchmarks and real-world robot deployments demonstrate that SER-Nav improves success rate and path efficiency, reduces redundant exploration, and transfers effectively from simulation to practical indoor environments. Overall, SER-Nav provides a lightweight, interpretable, and training-free solution for robust open-vocabulary embodied object navigation.




\bibliography{aaai2027}


\end{document}